\documentclass[10pt,twocolumn,letterpaper]{article}

\usepackage{iccv}
\usepackage{times}
\usepackage{epsfig}
\usepackage{graphicx}
\usepackage{amsmath}
\usepackage{amssymb}
\usepackage{multirow}

\usepackage[breaklinks=true,bookmarks=false]{hyperref}

\iccvfinalcopy 

\def\iccvPaperID{5} 

\ificcvfinal\fi

\begin{document}

\title{MOD: A Deep Mixture Model with Online Knowledge Distillation for Large Scale Video Temporal Concept Localization}

\author{Rongcheng Lin, Jing Xiao, Jianping Fan\\
University of North Carolina at Charlotte\\
{\tt\small \{rlin4, xiao, jfan\}@uncc.edu}
}

\maketitle
\ificcvfinal\thispagestyle{empty}\fi

\begin{abstract}
In this paper, we present and discuss a deep mixture model with online knowledge distillation (MOD) for large-scale video temporal concept localization, which is ranked 3rd in the 3rd YouTube-8M Video Understanding Challenge. Specifically, we find that by enabling knowledge sharing with online distillation, fintuning a mixture model on a smaller dataset can achieve better evaluation performance. Based on this observation, in our final solution, we trained and fintuned 12 NeXtVLAD models in parallel with a 2-layer online distillation structure. The experimental results show that the proposed distillation structure can effectively avoid overfitting and shows superior generalization performance. The code is publicly available at: \url{https://github.com/linrongc/solution_youtube8m_v3}
\end{abstract}

\section{Introduction}
Temporal concept localization within videos, which aims at automatically recognizing/retrieving topic related video segments, is one of critical and challenging problems to enable real world applications, including video search, video summarization, action recognition and video content safety etc. To accelerate the pace of research in this area, Google Research launched the 3rd YouTube-8M video understanding challenge and released 237K human-verified segment labels in addition to the about 6M noisy video-level labels. The goal is to retrieve related video segments from an unlabeled testing set for each of the 1000 classes. How to effectively leverage the large but noisy video-level labels for temporal localization is the main challenge.

One of the straightforward ideas is to pretrain models on the video-level dataset and then finetune the models using the smaller segment-level dataset. This approach turns out to be very effective in solving the problem. Also, we find that increasing parameter number of models by making the model wider can further improve the performance. But the marginal gains quickly diminish as the model are more likely to overfit the training dataset. Another way to increase the complexity of prediction system is to combine multiple models. Techniques to combine a set of weaker learners to create a strong learner, including bagging and boosting, are widely used in solving traditional machine learning problems. It is capable of reducing model variance and avoiding overfitting. However, in the era of deep learning, with millions even billions of parameters, single neural network could easily overfit the whole training dataset. The marginal gains from naive ensemble of multiple similar models also quickly diminish.

In this work, we propose a new approach by training a mixture of multiple base models in parallel with online knowledge distillation. With similar parameter number, a mixture model with online knowledge distillation can generalize better in the finetuning task than the wider model or the naive mixture. One possible explanation is that the online distillation part give each of the base models a holistic view of the similarity space and avoid the mixture model to overfit the smaller dataset. Based on this assumption, we built a 2-layer mixture model, which is a mixture of 4 MixNeXtVLAD models. And each of the MixNeXtVLAD model is a mixture of 3 base NeXtVLAD models\cite{DBLP:journals/corr/abs-1811-05014}. In summary, we trained 12 NeXtVLAD models in parallel and enabled a 2-layer online distillation structure. Experimental results show the superior generalization performance in the finetuning task, compared to simple mixture models or wider models.
\begin{figure*}
\centering
\includegraphics[width=0.8\linewidth]{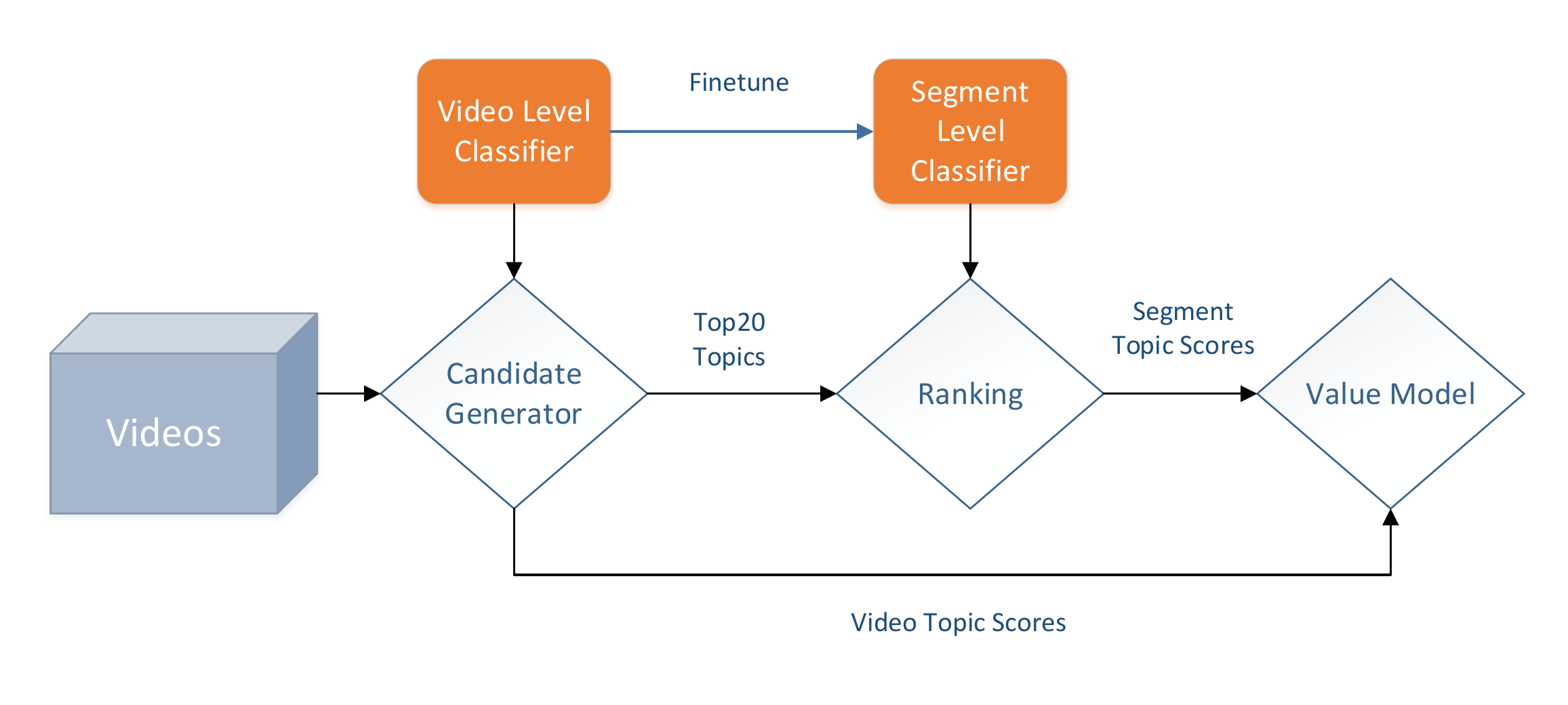}
\caption{Solution overview.}
\label{fig:system}
\end{figure*}
\section{Related Work}
\subsection{Deep Neural Network for Video Classification}
With the availability of large-scale video dataset, researchers proposed many deep neural networks and achieved remarkable advances in the field of video classification. In general, these approaches can be roughly summarized into 4 categories:
(a) \textbf{Spatiotemporal network}\cite{KarpathyCVPR14}\cite{Ji:2013:CNN:2412386.2412939}\cite{Tran:2015:LSF:2919332.2919929}. By regarding the temporal dimension as the extension of spatial dimensions, these models mainly rely on 2D or 3D convolution and pooling to aggregate information in the videos.
(b) \textbf{Recurrent network}\cite{Baccouche:2011:SDL:2177908.2177914}\cite{DBLP:journals/corr/BallasYPC15}. Apply recurrent neural networks, such as LSTM and GRU to aggregate the sequetial actions in the videos.
(c) \textbf{Two Stream Network}\cite{Simonyan:2014:TCN:2968826.2968890}\cite{DBLP:journals/corr/FeichtenhoferPZ16}\cite{DBLP:journals/corr/WuJWYXW15}\cite{43793}. Utilize optical flow images or similar features to model the motion in the video separately. The features extracted from frame images network and the optical flow network are fused to represent the videos.
(d) \textbf{Other approaches}\cite{Fernando_2015_CVPR}\cite{Wang_Transformation}\cite{DBLP:journals/corr/BilenFGV16}\cite{46699}. Use other information or methods to generate features for video representation and classification.
\subsection{Learnable Pooling Methods}
In the field of computer vision, aggregating multiple features into a single compact feature vector has been a long-standing research problem. Techniques, including BoW(Bag of visual Words)\cite{Sivic03}, FV(Fisher Vector)\cite{conf/cvpr/PerronninD07} and VLAD(Vector of Locally Aggregated Descriptors)\cite{conf/cvpr/JegouDSP10}, are widely used in computer vision systems\cite{Laptev08learningrealistic}\cite{Schuldt:2004:RHA:1018429.1020906}, including image/video retrieval, classification and localization. Recently, inspired by the work of VLAD, a learnable pooling network, NetVLAD, is firstly introduced in \cite{Arandjelovic16} to solve the problem of place recognition. In the task of video understanding with pre-extracted frame level features, NetVLAD shows superior performance\cite{Arandjelovic16}\cite{kali2018BuildingAS}. Several other variants, including NeXtVLAD\cite{DBLP:journals/corr/abs-1811-05014} and Non-local NetVLAD\cite{DBLP:journals/corr/abs-1810-00207} etc, were proposed to further improve the parameter efficiency and generalization performance.
\subsection{Knowledge Distillation}
Knowledge distillation\cite{44873} is an effective and popular approach for model compression by distilling a complex teacher model to a simpler student model. The success of transferring the dark knowledge between networks has inspired many novel research work in computer vision\cite{DBLP:journals/corr/LiH16e}\cite{DBLP:journals/corr/RusuRDSKKPH16}\cite{DBLP:journals/corr/ChenGS15}. Recently, researchers find that, rather than the one-way knowledge transfer, enabling collaborative learning of several simple student models with a two-way knowledge sharing can achieve superior results\cite{DBLP:journals/corr/ZhangXHL17}\cite{on_the_fly} and can be efficiently trained within a distributed training system\cite{DBLP:journals/corr/abs-1804-03235}.
\section{Solution}
\subsection{Solution Overview}

The overall structure of the solution to generate video segments for each of the 1000 topics is illustrated in Figure \ref{fig:system}. The system is comprised of three phases: (1) candidate generation via a video level classifier. Only the top 20 topics are considered to be existed in the video. An offline analysis demonstrate that those candidates cover over 97\% of the positive samples(recall) in the segment training dataset. This step significantly reduce the search space. (2) a segment level classifier is used as a ranker to assign probabilities to each of the 5s segments in the video. The segment level classifier is directly finetuned from the video level classifier. (3)we combine the video topic scores $P_{vid}(K)$and segment topic score $P_{seg}(K)$ via a value model:
\begin{equation}
P(K=k) = P_{vid}^{0.05}(K=k) * P_{seg}^{0.95}(K=k)
\end{equation}
And finally, for each of the 1000 topics, we retrieve the top 10K video segments ranked by the combined score.
The whole system relied heavily on the performance of video and segment level classifier. How to build a accurate and robust classifier is the essential part of the solution.
\subsection{NeXtVLAD Model}
The base model used in our classifier is NeXtVLAD model, which achieved the best single model performance in the 2nd YouTue-8M video understanding challenge. Before diving into the final solution, we will first briefly review the NeXtVLAD pooling network and NeXtVLAD model for video classification.
\begin{figure}
\centering
\includegraphics[width=1.0\linewidth]{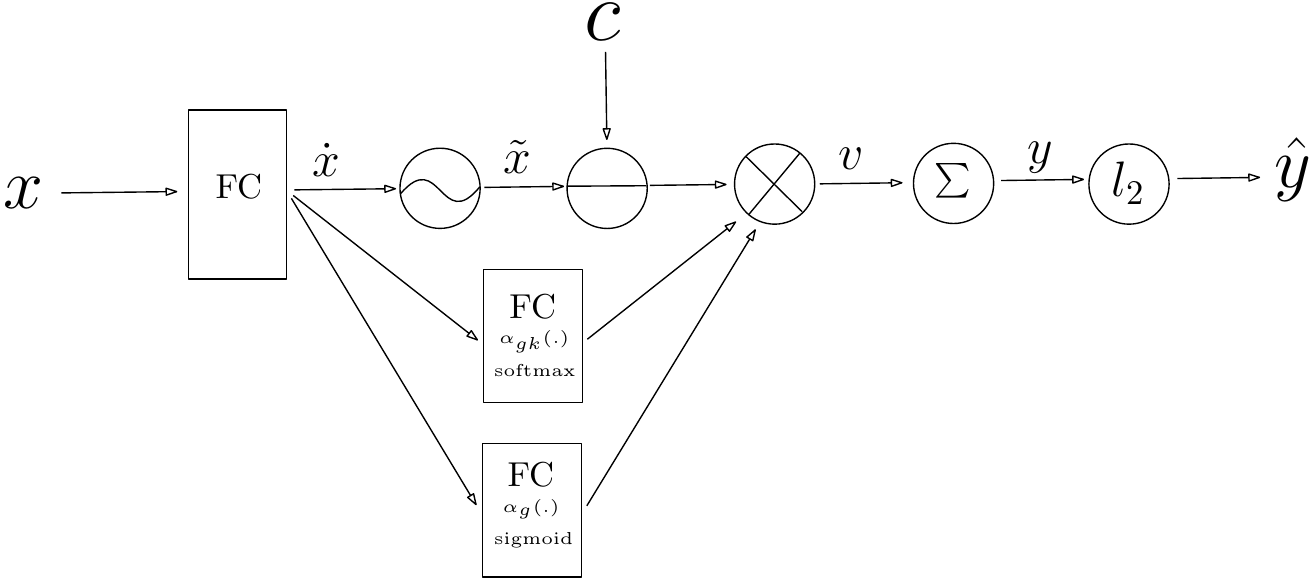}
\caption{NeXtVLAD pooling network.}
\label{fig:nextvlad}
\end{figure}

\begin{figure*}
\centering
\includegraphics[width=1.0\linewidth]{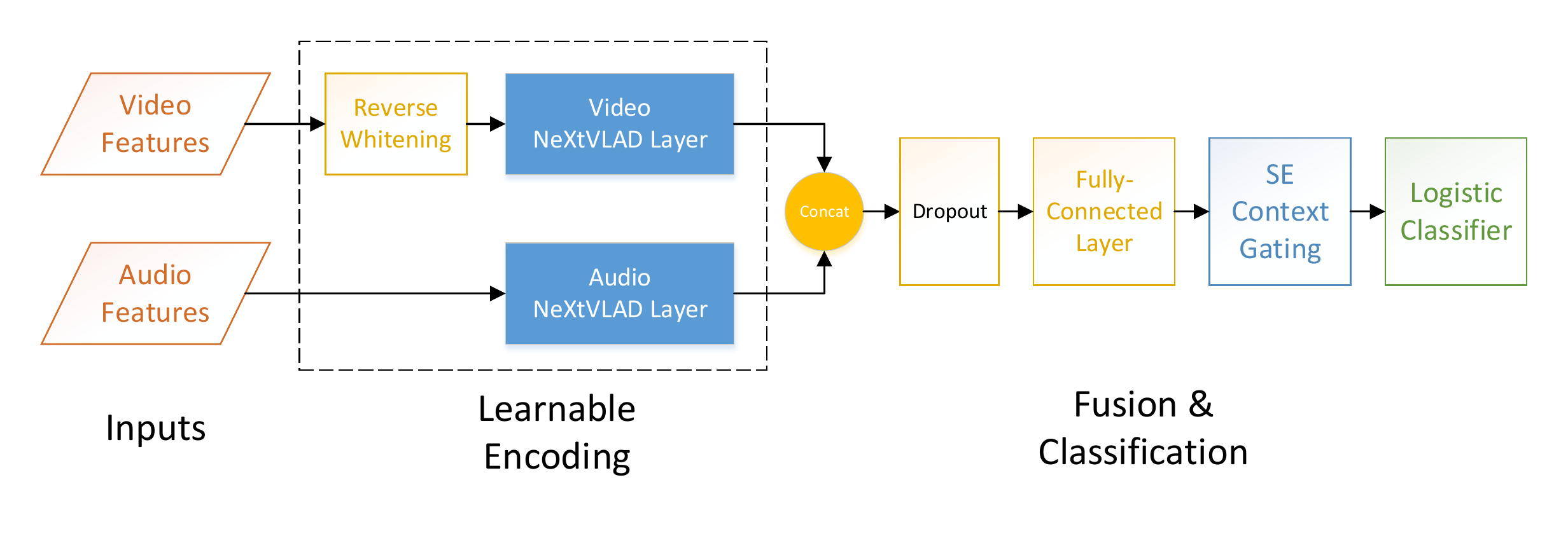}
\caption{Overview of a NeXtVLAD model designed for Youtube-8M video classification.}
\label{fig:nextvlad_network}
\end{figure*}

A NeXtVLAD pooling network, as shown in Figure \ref{fig:nextvlad}, is a variant of NetVLAD, which is a differentiable network inspired by traditional Vector of Locally Aggregated Descriptors(VLAD).
Considering a video input $x$ with M frames and each of the frame is represented as N-dimension feature, a NeXtVLAD expand the input dimension by a factor of $\lambda$ at first via a linear projection to be $\dot{x}$ with a shape of $(M, \lambda N)$. Then $\dot{x}$ is splitted into $G$ groups, each of which is represented as $\tilde{x}^g$. The NeXtVLAD pooling is a mixture of group-level NetVLAD aggregations into $K$ clusters:
\begin{equation}
y_{jk} = \sum_g \alpha_g(\dot{x}_i) v^g_{jk}
\end{equation}
$$
\forall i \in \{1, ..., M\},j \in \{1, ..., N\},k \in \{1, ..., K\},g \in \{1, ..., G\}
$$
in which $\alpha_g(\dot{x}_i)$ is group level attention function:
\begin{equation}
\alpha_g(\dot{x}) = \sigma(w^T_g \dot{x} + b_g)
\end{equation}
and $v^g_{jk}$ is the output of group level NetVLAD aggregation:
\begin{equation}
v^g_{jk} = \sum_i \alpha_{gk}(\dot{x}_i)(\tilde{x}_{ij}^g - c_{kj})
\end{equation}
Finally, a $l_2$ normalization, a.k.a. intra-normalization, is applied to the aggregated features for each of the clusters:
\begin{equation}
\hat{y}_{jk} = \frac{y_{jk}}{\|y_{k}\|_2}
\end{equation}
The $l_2$ normalization is one of essential parts to make features extracted from different videos or video segments are comparable. And it is also one of the reasons why finetuning a video-level model can work well as a segment level classifier.
\begin{figure}
\centering
\includegraphics[width=1.0\linewidth]{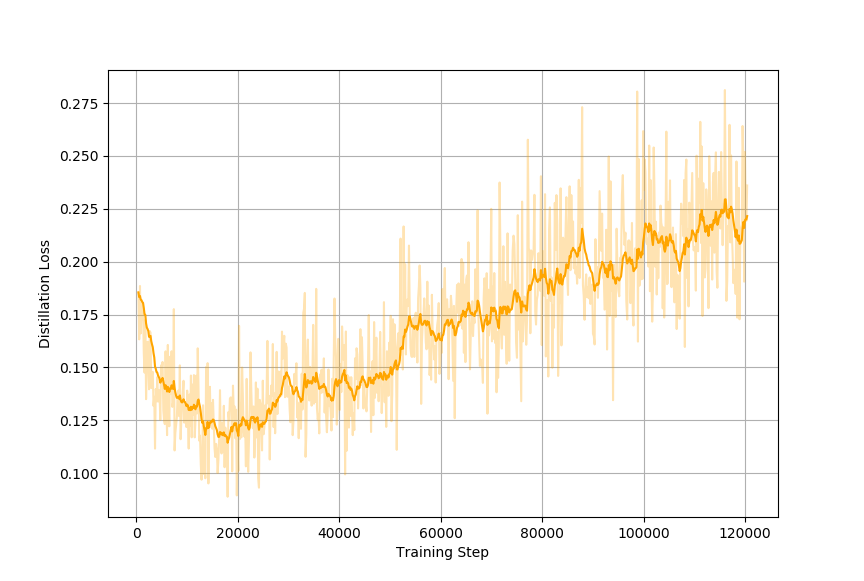}
\caption{An example of distillation loss of a MixNeXtVLAD Model.}
\label{fig:distill_loss}
\end{figure}
\begin{figure}
\centering
\includegraphics[width=0.9\linewidth]{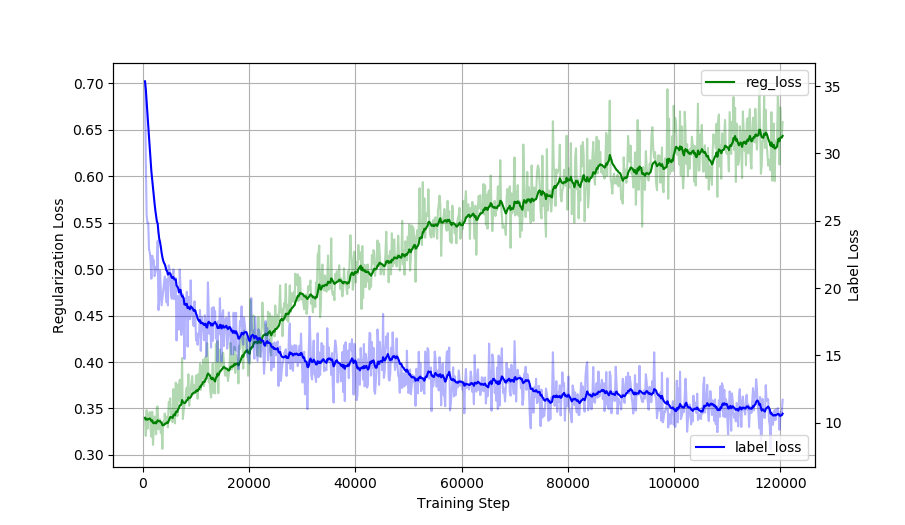}
\caption{An example of label loss and regularization loss of a MixNeXtVLAD Model.}
\label{fig:training}
\end{figure}
As illustrated in Figure \ref{fig:nextvlad_network}, in the NeXtVLAD model designed for video classification, video and audio features are aggregated by two NeXtVLAD pooling networks separately. Then the aggregated features are concatenated and fed into a dropout layer before a FC layer is applied to reduce the dimension of the encoded features. After the dropout layer, a context gating layer is appended to capture the dependency among topics. Finally, a logistic model is used as the final classifier.
\subsection{MixNeXtVLAD Model}
\begin{figure*}
\centering
\includegraphics[width=0.8\linewidth]{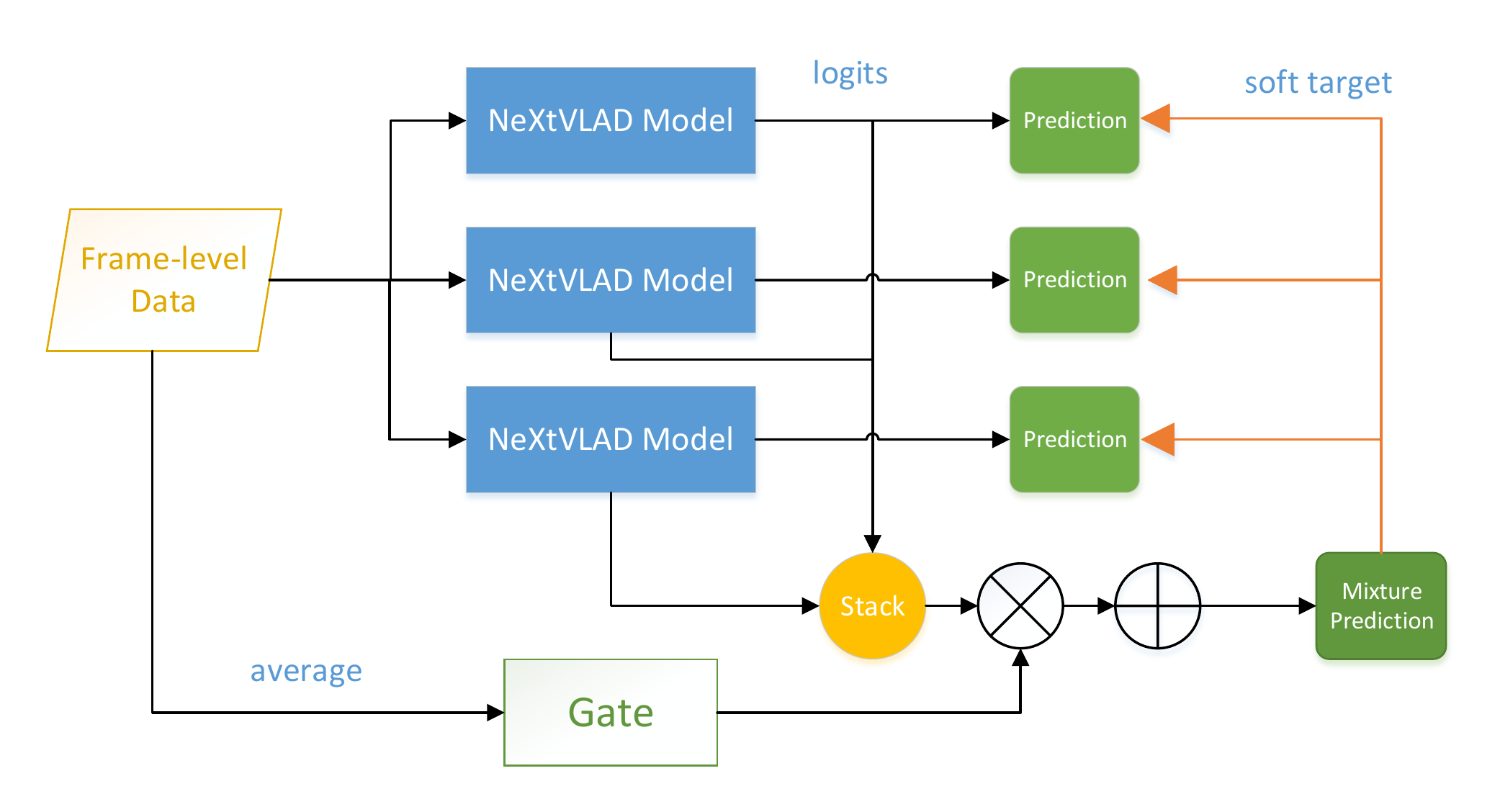}
\caption{Overview of a mixture of 3 NeXtVLAD models(MixNeXtVLAD) with online knowledge distillation. The orange arrows indicate the distillation of knowledge from the mixture prediction to the predictions of sub-models.}
\label{fig:mix_nextvlad}
\end{figure*}

Training multiple base models in parallel and distill knowledge from the mixture predictions to sub-models via a distillation loss is firstly introduced in \cite{on_the_fly} and applied to the video classification problem in \cite{DBLP:journals/corr/abs-1811-05014}. The MixNeXtVLAD model is a mixture of 3 NeXtVLAD model with on-the-fly knowledge distillation. As shown in Figure \ref{fig:mix_nextvlad}, the logit $z^e$ of mixture prediction $p^e$ is the weighted sum of the logits $z^m$ from predictions $p^m$ of sub-models. Given the ground truth label $y$, The final loss of the MixNeXtVLAD model is:
\begin{equation}
\begin{gathered}
\mathcal{L} = \sum_{m=1}^3 \mathcal{L}_{bce}(y, p_m) + \mathcal{L}_{bce}(y, p_e) \\
+ T^2 * \sum_{m=1}^3 \mathcal{KL}(Soft(p_e, T)\|Soft(p_m, T))
\end{gathered}
\end{equation}
where $L_{bce}$ is the binary cross entropy and $\mathcal{KL}(Soft(p_e, T)\|Soft(p_m, T))$ represents distillation loss, which is the KL divergence between the soften predictions:
\begin{equation}
Soft(p, T) = Softmax(z/T)
\end{equation}
in which $z$ is the logits of prediction $p$. A larger T value will emphasize more on the smaller values in the prediction and thus share more knowledge about the learned similarity in the task space.

One of the main assumptions is that the online distillation loss will provide a holistic view for sub-models to the task space during training. If we dive closer to the binary cross entropy loss of the mixture prediction $\mathcal{L}_{bce}(y, p_e)$, we can find the loss capture the remaining part which is not covered by the predictions from all the sub-models. In other word, if one sub-model capture part of the true prediction, then the information will be ignored by the loss for other sub-models. As a result, the predictions of sub-models are diversified during training. The distillation loss between mixture prediction and individual prediction will ensure the sub-models have the holistic view of the whole task space. Figure \ref{fig:distill_loss} shows one example of the online distillation loss of a MixNeXtVLAD Model during training. The distillation loss is optimized(decreasing) at the beginning then increasing steadily as we further minimize the whole objective function (Figure \ref{fig:training}). The increase of the distillation loss at later stage of training is a implicit proof of our assumption.

\subsection{Deep Mixture of NeXtVLAD Models with Online Distillation}
\begin{figure*}
\centering
\includegraphics[width=0.8\linewidth]{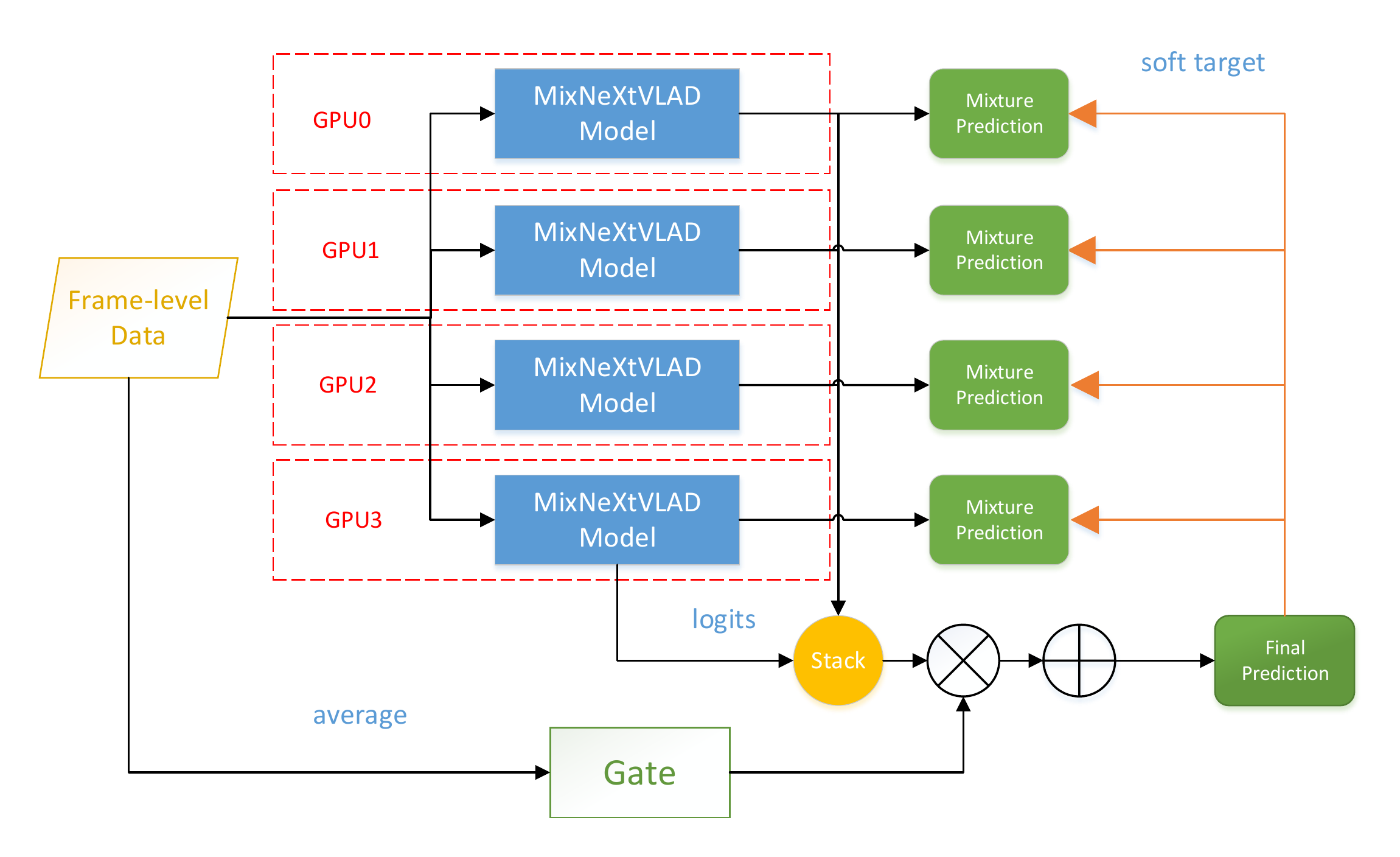}
\caption{Overview of a mixture of 4 MixNeXtVLAD models with online knowledge distillation. Each of the MixNeXtVLAD model is a mixture of 3 base NeXtVLAD model. The orange arrows indicate the distillation of knowledge from the final prediction to mixture predictions of subordinated mixture models. Each of the 4 MixNeXtVLAD model is located in the separate GPU.}
\label{fig:mix_mix}
\end{figure*}

A deep mixture of NeXtVLAD models with online distillation (MODNeXtVLAD thereafter), which is the model used as our final solution, is a intuitive extension of the MixNeXtVLAD Model. As shown in Figure \ref{fig:mix_mix}, MODNeXtVLAD is a mixture of 4 MixNeXtVLAD models, each of which is a mixture of 3 base NeXtVLAD Models. So in total, in MODNeXtVLAD, 12 NeXtVLAD models are trained and finetuned simultaneously. As for the knowledge distillation part, knowledge is firstly distilled from the final prediction to each of the mixture models, then from mixture prediction to each of the NeXtVLAD models. For simplicity, we apply the same parameter(T in this case) in the two-stage knowledge distillation.

To be general, the MOD structure forms a simple 2-layer model-level hierarchy, where each sub-tree is an independent mixture model and knowledge is distilled from root to leaves one layer at a time. One advantage of the MOD structure is its suitability for distributed training. Except for knowledge distillation loss and mixture of logits, models in different subtrees can be trained independently and thus can be located in different physical devices and the communication(network) overhead is negligible. So in our implementation, we applied model parallel distributed training strategy instead of data parallel to improve the training efficiency. With data parallel strategy, the training speed of one NeXtVLAD model in 2 Nvidia 1080TI GPUs is about 400 examples per second. By enabling model parallel in training 12 same NeXtVLAD models with MOD structure and model parallel strategy, we can achieve a training speed of 140+ examples per second using 4 Nvidia 1080 TI GPUs.

\begin{table*}
\caption{Performance comparison of single models trained on frame-level features. The parameters inside the parenthesis represents (group number G, cluster number K, expansion factor $\lambda$, hidden size H).}\label{tbl:single_model}
  \centering
  \begin{tabular}{l|lccc}
    \hline
    &Model & Parameter & Private LB & Public LB \\
    \hline
    Without Finetune&Dummy prediction & 83M & 0.64809 & 0.66188\\
    \hline \hline
    \multirow{2}{*}{Without Pretrain}&DBoF Baseline & 16M & 0.69882 & 0.71077 \\
    &NeXtVLAD(8G, 128K, X2, 2048H) & 83M & 0.77009 & 0.77730\\
    \hline\hline
    \multirow{3}{*}{Pretrain and finetune}&NeXtVLAD(8G, 128K, X2, 2048H) & 83M & 0.79642 & 0.80635\\
    &NeXtVLAD\_large(8G, 256K, X4, 2048H)& 320M & 0.80586 & 0.81611\\
    &NeXtVLAD\_distill(8G, 128K, X2, 2048H) & 83M & 0.81509 & 0.82267\\
    \hline
  \end{tabular}\\
\end{table*}
\begin{table*}
\caption{Performance comparison of mixture models. All the base models used are NeXtVLAD(8G, 128K, X2, 2048) except for MixNeXtVLAD\_large, which take NeXtVLAD\_large as the base model. }\label{tbl:mix_model}
  \centering
  \begin{tabular}{l|lcccc}
    \hline
    &Model & Base Model Number & Parameter & Private LB & Public LB \\
    \hline
    \multirow{4}{*}{One-Layer Mixture}&MixNeXtVLAD(T=0) & 3 & 250M & 0.80797 & 0.81688\\
    &MixNeXtVLAD(T=1) & 3 & 250M & 0.81125 & 0.82023 \\
    &MixNeXtVLAD(T=10) & 3 & 250M & 0.81617 & 0.82477 \\
    &MixNeXtVLAD(T=20) & 3 & 250M & 0.81984 & 0.82699 \\
    &MixNeXtVLAD\_large(T=20) & 4 & 1280M & 0.82262 & 0.83014\\
    \hline \hline
    Two-Layer Mixture &MODNeXtVLAD(T=20) & 12 & 1000M & 0.82512 & 0.83251\\
    \hline
  \end{tabular}\\
\end{table*}
\section{Experimental Results}
\subsection{Dataset and Evaluation Metrics}
\subsubsection{YouTuebe-8M Video Dataset}
Youtube-8M video dataset\cite{DBLP:journals/corr/Abu-El-HaijaKLN16} consists of 6.1M popular videos from Youtube.com. These videos are splitted in to 3 partitions: training(70\%), validation(20\%) and testing(10\%).
For each video in the training and validation dataset, one or multiple labels(3.0 labels/video on average) are generated by an annotation machine from a vocabulary of 3862 visual entities. These video-level labels is not verified by human and thus noisy in terms of label quality.
For every second of the videos, frame-level features, including a 1024-dimensional visual feature and a 128-dimensional audio feature, are precomputed and provided for model training.
\subsubsection{Youtube-8M Segment Dataset}
As an extension of the original YouTube-8M dataset, the segement dataset contains 237K human-verified segment labels on 1000 classes. These segments are sampled from the validation set of the Youtube-8M video dataset and contains exactly 5 frames. Each segment label indicates whether the 5s segments contains objects of the target class. Compared to the video dataset, this segment dataset is clean but much smaller. How to leverage the large amount but noisy video level labels is one of the main challenges.
\subsubsection{Evaluation Metrics}
In the 3rd Youtube-8M video understanding challendge, submissions are evaluated using Mean Average Precision at K (MAP@K):
\begin{equation}
MAP@K = \frac{1}{C} \sum_{c=1}^C \frac{\sum_{k=1}^KP(k)*rel(k)}{N_c}
\end{equation}
where $C$ is the number of classes, $N_c$ is the total positive samples in the class, $P(k)$ is the precision at cutoff $k$ and $rel(k)$ is an indicator function to represent whether the $k_{th}$ items belong to class $c$. The metric is an approximate of the area under Precision-Recall curve.
\subsection{Implementation Details}
Our implementation is based on the TensorFlow starter code provided by the organizer. All the models are run at a machine with 4 Nvidia GPUs. We follows the same settings in \cite{} to train video level models. For a fair comparison, each model is trained for about 500K steps to guarantee the convergence. As for larger models, including our final model, we use a batch size of 80 to avoid out of memory in GPUs.

In the finetuning stage, all the models are trained with a batch size of 512. The dropout rate and the l2-normalization penalty are increased to 0.75 and 1e-4 respectively aiming to prevent overfitting. Models are trained for 10 epochs on the segment dataset using the Adam optimizer with a intial learning rate of 0.0002. The learning rate is decayed by a factor of 0.8 for every 1M examples. More training details can be found at \url{https://github.com/linrongc/solution_youtube8m_v3}

\subsection{Model Evaluation}

\subsubsection{Single Model Comparison}
The performance and parameter number of single models are summarized in Table \ref{tbl:single_model}. The evaluation metrics presented in the table is MAP@100000. The models included in the comparison are:
\begin{itemize}
\item[-] Dummy prediction. A NeXtVLAD model trained only using the video level labels. All the segments in the video are considered to contains the same content.
\item[-] DBoF Baseline. A deep bag of frame model provided in the starter code with 2048 clusters and a hidden size of 1024. The final classifier is a MOE(mixture of experts) model with 5 experts.
\item[-] NeXtVLAD. The best single model in the 2nd YouTube-8M video understanding challenge.
\item[-] NeXtVLAD\_distill. One single NeXtVLAD model used in the two-layer mixture model with online knowledge distillation. It is trained with other 11 NeXtVLAD models with the same settings. But in the inference stage, those 11 NeXtVLAD models are removed.
\end{itemize}

Generally speaking, models which are pretrained on the larger video dataset outperform models without pretrain. While a larger NeXtVLAD model with more parameters can achieve better MAP score, one single and small NeXtVLAD model used in 2-layer mixture with online knowledge distillation shows superior performance.

\subsubsection{Mixture Model Comparison}
We evaluate the one-layer MixNeXtVLAD model (Figure \ref{fig:mix_nextvlad}) with different settings and one two-layer mixture of NeXtVLAD model, which is the model used in our final submission. As illustrated in Table \ref{tbl:mix_model}, the MixNeXtVLAD model without knowledge distillation(T=0) shows the similar performance with the larger NeXtVLAD model(NeXtVLAD\_large). By gradually increasing the value of T, the generalization performance is improved accordingly. The results indicates that, with higher value of temperature(T), more knowledge are distilled from the mixture model to each single model. The knowledge distillation part can effectively avoid model overfitting.

Also, a two-layer mixture model, MODNeXtVLAD, can easily outperform one-layer mixture model even with less number of parameters. The results directly prove the parameter efficiency and better generalization performance of the proposed deep mixture structure with online distillation.

\section{Conclusion and Future Work}
In this paper, we proposed a novel deep mixture model with online knowledge distillation and evaluated the model performance in the 3rd YouTube-8M video understanding challenge. The model can be efficiently trained in a distributed training system because of the low communication cost between the base models. The experimental results shows that, in a finetune task, online knowledge distillation can effectively improve the generalization performance of the mixture model.

Due to the resource limit, only a 2-layer mixture model with online distillation is included in the experiment. Whether a deeper mixture model with online knowledge distillation can further improve the generalization performance still need to be verified.
{\small
\bibliographystyle{ieee_fullname}
\bibliography{egbib}
}

\end{document}